\documentclass[manuscript,screen]{acmart}

\AtBeginDocument{%
  \providecommand\BibTeX{{%
    \normalfont B\kern-0.5em{\scshape i\kern-0.25em b}\kern-0.8em\TeX}}}

\setcopyright{acmcopyright}
\copyrightyear{2020}
\acmYear{2020}
\acmDOI{10.1145/1122445.1122456}

\usepackage{mdframed,lipsum}
\usepackage{subcaption}

\begin{document}

\title{Multilingual Offensive Language Identification for Low-resource Languages}

\author{Tharindu Ranasinghe}
\affiliation{%
  \institution{University of Wolverhampton}
  \streetaddress{Wulfruna Streen}
 \city{Wolverhampton}
  \country{UK}
 \postcode{WV1 1LY}
}
\email{T.D.RanasingheHettiarachchige@wlv.ac.uk}

\author{Marcos Zampieri}
\affiliation{%
  \institution{Rochester Institute of Technology}
  \streetaddress{92 Lomb Memorial Drive}
  \city{Rochester}
\state{NY}
 \country{USA}
  \postcode{14620}
}
\email{marcos.zampieri@rit.edu}

\renewcommand{\shortauthors}{Ranasinghe and Zampieri}

\begin{abstract}
  Offensive content is pervasive in social media and a reason for concern to companies and government organizations. Several studies have been recently published investigating methods to detect the various forms of such content (e.g. hate speech, cyberbulling, and cyberaggression). The clear majority of these studies deal with English partially because most annotated datasets available contain English data. In this paper, we take advantage of available English datasets by applying cross-lingual contextual word embeddings and transfer learning to make predictions in low-resource languages. We project predictions on comparable data in Arabic, Bengali, Danish, Greek, Hindi, Spanish, and Turkish. We report results of 0.8415 F1 macro for Bengali in TRAC-2 shared task \cite{trac2020}, 0.8532 F1 macro for Danish and 0.8701 F1 macro for Greek in OffensEval 2020 \cite{zampieri-etal-2020-semeval}, 0.8568 F1 macro for Hindi in HASOC 2019 shared task \cite{hasoc2019} and 0.7513 F1 macro for Spanish in in SemEval-2019 Task 5 (HatEval) \cite{hateval2019} showing that our approach compares favourably to the best systems submitted to recent shared tasks on these three languages. Additionally, we report competitive performance on Arabic, and Turkish using the training and development sets of OffensEval 2020 shared task. The results for all languages confirm the robustness of cross-lingual contextual embeddings and transfer learning for this task.
\end{abstract}



\maketitle

\section{Introduction}

Offensive posts on social media result in a number of undesired consequences to users. They have been investigated as triggers of suicide attempts and ideation, and mental health problems \cite{bonanno2013cyber,bannink2014cyber}. One of the most common ways to cope with offensive content online is training systems capable of recognizing offensive messages or posts. Once recognized, such offensive content can be set aside for human moderation or deleted from the respective platform (e.g. Facebook, Twitter) preventing harm to users and controlling the spread of abusive behavior in social media. 

There have been several recent studies published on this topic focusing on different aspects of offensiveness such as abuse \cite{mubarak2017}, aggression \cite{trac2018report,trac2020}, cyber-bullying \cite{rosa2019automatic}, and hate speech \cite{malmasi2017,malmasi2018,rottger2020hatecheck} to name a few. While there are a few studies published on languages such as Arabic \cite{mubarak2020arabic} and Greek \cite{pitenis2020}, most studies and datasets created thus far have focused on English. Data augmentation \cite{ghadery2020liir} and multilingual word embeddings \cite{pamungkas2019cross} have been applied to take advantage of existing English datasets to improve the performance in systems dealing with languages other than English. To the best of our knowledge, however, state-of-the-art cross-lingual contextual embeddings such as XLM-R \cite{conneau2019unsupervised} have not yet been applied to offensive language identification. To address this gap, we evaluate the performance of cross-lingual contextual embeddings and transfer learning (TL) methods in projecting predictions from English to other languages. We show that our methods compare favourably to state-of-the-art approaches submitted to recent shared tasks on all datasets. The main contributions of this paper are the following:

\begin{enumerate}
    \item We apply cross-lingual contextual word embeddings to offensive language identification. We take advantage of existing English data to project predictions in seven other languages: Arabic, Bengali, Danish, Greek, Hindi, Spanish, and Turkish.
    \item We tackle both off-domain and off-task data for Bengali. We show that not only these methods can project predictions on different languages but also on different domains (e.g. Twitter vs. Facebook) and tasks (e.g. binary vs. three-way classification).
    \item We provide important resources to the community: the code, and the English model will be freely available to everyone interested in working on low-resource languages using the same methodology.
\end{enumerate}

\noindent Finally, it is worth noting that the seven languages other than English included in this study are all languages with millions of speakers. Their situation in terms of language resources is more favorable and not comparable to the situation of minority or endangered languages for which there are virtually no datasets available, often called truly low-resource languages \cite{agic2016multilingual} or extremely low-resource languages \cite{tapo2020neural}. That said, we consider these seven languages as low-resourced in the context of offensive language identification due to the lack of large available offensive language datasets in these languages especially when compared to English. The methods presented in this paper can be used to truly low-resource languages addressing data scarcity. 

\section{Related Work}

There have been different types of abusive content addressed in recent studies including hate speech \cite{malmasi2018}, aggression \cite{trac2018report,trac2020}, and cyberbullying \cite{rosa2019automatic}. A few annotation taxonomies, such as the one proposed by OLID \cite{OLID} and replicated in other studies \cite{SOLID}, try to take advantage of the similarities between these sub-tasks allowing us to consider multiple types of abusive language at once. 

In terms of computational approaches, a variety of methods have been tested to identify abusive posts automatically including simple profanity lexicons, machine learning classifiers like Naive Bayes and SVMs, and deep learning representations and neural networks like convolutional neural networks (CNN), recurrent neural networks (RNN) \cite{hettiarachchi-ranasinghe-2019-emoji}, LSTM \cite{aroyehun2018aggression,majumder2018filtering} and transformers \cite{ranasinghe2019brums}. In this section we briefly summarise related work addressing the most common abusive language phenomena.

\vspace{3mm}

\noindent {\bf Aggression identification:} The two editions of the TRAC shared task on Aggression Identification organized in 2018 and 2020 \cite{trac2018report,trac2020} provided participants with annotated Facebook dataset containing posts and comments labeled as non-aggressive, covertly aggressive, and overtly aggressive. The best-performing systems in the 2018 edition competition used deep learning approaches based on CNN, RNN, and LSTM  \cite{aroyehun2018aggression,majumder2018filtering}. In Section \ref{sec:results} we compare our results for Bengali with the best performing methods of TRAC 2020. 

\vspace{3mm}

\noindent {\bf Cyberbullying detection:} Cyberbulling is another popular topic with several studies published such as \cite{xu2012learning} who used sentiment analysis methods and topic modelling and \cite{dadvar2013improving} who addressed the problem using profiling-related features like the frequency of curse words in users' messages.

\vspace{3mm}

\noindent \textbf{Hate Speech:} Hate speech is by far the most studied phenomenon with several studies published on various languages \cite{burnap2015cyber,djuric2015hate}. The recent HatEval \cite{hateval2019} competition at SemEval 2019 addressed hate speech against women and migrants featuring datasets in English and Spanish. 

\vspace{3mm}

\noindent \textbf{Offensive Language:} In addition to the popular OffensEval shared task, which introduced a taxonomy encompassing multiple types of abusive and offensive content discussed in more detail in Section \ref{offenseval}, there have been several studies published on identifying offensive posts \cite{pitenis2020}. The GermEval \cite{wiegand2018overview} shared task provided participants with a dataset containing 8,500 annotated German tweets. Two sub-tasks were organized, the first sub-task was to discriminate between offensive and non-offensive tweets and the second sub-task was to discriminate between profanity, insult, and abuse. 

\vspace{3mm}

\noindent \textbf{Other Languages:} While the clear majority of studies deal with English, there have been a number of studies dealing of other languages as well. Examples included Dutch \cite{vanhee2015,tulkens2016a}, German \cite{ross}, Italian \cite{Pelosi2017MiningOL}, and the languages studied in the present paper. 

\section{Data}
\label{sec:data}

To carry out the experiments presented in this paper, we acquire datasets in English and other seven languages: Arabic, Bengali, Danish, Greek, Hindi, Spanish, and Turkish listed in Table \ref{tab:data}. 

\begin{table}[!ht]
\centering
\setlength{\tabcolsep}{4.5pt}
\begin{tabular}{lccp{7.0cm}}
\hline
\bf Lang. & \bf Instances & \bf Source & \bf Labels  \\ \hline

Arabic & 8,000 & T & offensive, non-offensive \\
Bengali & 4,000 & Y & overtly aggressive, covertly aggressive, non aggressive     \\
Danish & 2,961 & F, R & offensive, non-offensive \\
English & 14,100 & T & offensive, non-offensive \\
Greek & 8,743 & T & offensive, non-offensive \\
Hindi & 8,000 & T & hate offensive, non hate-offensive \\
Spanish & 6,600 & T & hateful, non-hateful      \\
Turkish & 31,756 & T & offensive, non-offensive \\
\hline
\end{tabular}
\caption{Instances (Inst.), source (S) and labels in all datasets. F stands for Facebook, R for Reddit, T for Twitter and Y for Youtube.}
\label{tab:data}
\end{table}

\noindent The Bengali, Hindi, and Spanish datasets have been used in shared tasks in 2019 and 2020 allowing us to compare the performance of our methods to other approaches. The Hindi dataset \cite{hasoc2019} was used in the HASOC 2019 shared task while the Spanish dataset \cite{hateval2019} was used in SemEval-2019 Task 5 (HatEval). They both contain Twitter data and two labels. The Bengali dataset \cite{trac2-dataset} was used in the TRAC-2 shared task \cite{trac2020} on aggression identification. It is different than the other three datasets in terms of domain (Youtube instead of Twitter) and set of labels (three classes instead of binary) allowing us to compare the performance of cross-lingual embeddings on off-domain data and off-task data. 

The Arabic, Danish, Greek, and Turkish datasets are official datasets of the OffensEval 2020 competition hosted at SemEval 2020 (Task 12) \cite{zampieri-etal-2020-semeval}. We report the results, we obtained for Danish and Greek test sets. in Section \ref{offenseval}. However, since Arabic and Turkish are morphologically rich, they require language specific segmentation approaches which we hope to conduct in the future. Therefore we only report results on the development set in Section \ref{offenseval} and a detailed explanation of future work on Arabic and Turkish is reported in Section \ref{sec:future}.

Finally, as our English dataset, we choose the Offensive Language Identification Dataset (OLID) \cite{OLID}, used in the SemEval-2019 Task 6 (OffensEval) \cite{offenseval}. OLID is arguably one of the most popular offensive language datasets. It contains manually annotated tweets with the following three-level taxonomy and labels:


\begin{itemize}
    \item[\bf A:] Offensive language identification - offensive vs. non-offensive;
    \item[\bf B:] Categorization of offensive language - targeted insult or thread vs. untargeted profanity;
    \item[\bf C:] Offensive language target identification - individual vs. group vs. other.
\end{itemize}


\noindent We chose OLID due to the flexibility provided by its hierarchical annotation model which considers multiple types of offensive content in a single taxonomy (e.g. targeted insults to a group are often {\em hate speech} whereas targeted insults to an individual are often {\em cyberbulling}). This allows us to map OLID level A (offensive vs. non-offensive) to labels in the other three datasets. OLID's annotation model is intended to serve as a general-purpose model for multiple {\em abusive language detection sub-tasks} as described by Waseem et al. (2017) \cite{waseem2017understanding}. The transfer learning strategy used in this paper provides us with the opportunity to evaluate how closely the OLID labels relate to the classes in other datasets which were annotated using different guidelines and definitions (e.g. {\em aggression} and {\em hate speech}).

\section{Methodology}

Transformer models have been used successfully for various NLP tasks \cite{devlin2019bert} such as text classification \cite{ranasinghe-hettiarachchi-2020-brums, hettiarachchi-ranasinghe-2020-infominer}, NER \cite{ranasinghemudes, taher-etal-2019-beheshti}, context similarity \cite{hettiarachchi-ranasinghe-2020-brums}, language identification \cite{jauhiainen2021comparing} etc. Most of the tasks were focused on English language due to the fact the most of the pre-trained transformer models were trained on English data. Even though, there were several multilingual models like BERT-m \cite{devlin2019bert} there were many speculations about its ability to represent all the languages \cite{pires-etal-2019-multilingual} and although BERT-m model showed some cross-lingual characteristics it has not been trained on crosslingual data \cite{karthikeyan2020cross}. The motivation behind this methodology was the recently released cross-lingual transformer models - XLM-R \cite{conneau2019unsupervised} which has been trained on 104 languages. The interesting fact about XLM-R is that it is very compatible in monolingual benchmarks while achieving best results in cross-lingual benchmarks at the same time \cite{conneau2019unsupervised}. The main idea of the methodology is that we train a classification model on a resource rich, typically English, using a cross-lingual transformer model and perform transfer learning on a less resource language. 

There are two main parts of the methodology. Subsection \ref{subsec:classification} describes the classification architecture we used for all the languages. In Subsection \ref{subsec:transfer} we describe the transfer learning strategies we used to utilise English offensive language data in predicting offense in less-resourced languages. 

\subsection{XLM-R for Text Classification}
\label{subsec:classification}
Similar to other transformer architectures XLM-R transformer architecture can also be used for text classification tasks \cite{conneau2019unsupervised}. XLM-R-large model contains approximately 125M parameters with 12-layers, 768 hidden-states, 3072 feed-forward hidden-states and 8-heads \cite{conneau2019unsupervised}. It takes an input of a sequence of no more than 512 tokens and outputs the representation of the sequence. The first token of the sequence is always [CLS] which contains
the special classification embedding \cite{10.1007/978-3-030-32381-3_16}.

For text classification tasks, XLM-R takes the final hidden state h of the first token [CLS] as the representation of the whole sequence. A simple softmax classifier is added to the top of XLM-R to predict the probability of label c: as shown in Equation \ref{equ:softmax} where W is the task-specific parameter matrix \cite{ranasinghe-hettiarachchi-2020-brums, hettiarachchi-ranasinghe-2020-infominer}.

\begin{equation}
\label{equ:softmax}
p(c|\textbf{h}) = softmax(W\textbf{h}) 
\end{equation}

\begin{figure}[ht]
\centering
\includegraphics[scale=0.4]{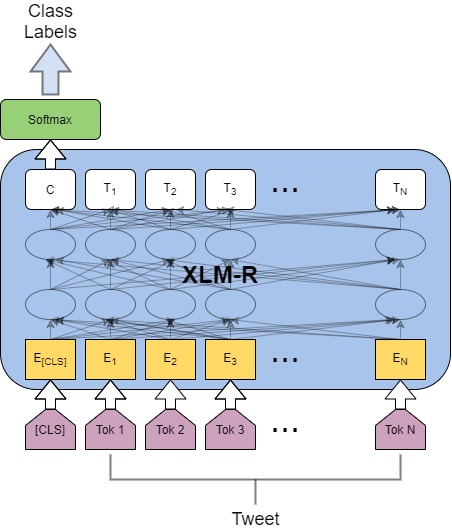}
\caption{Text Classification Architecture \cite{ranasinghe-zampieri-2020-multilingual} }
\label{fig:architecture}
\end{figure}

\noindent We fine-tune all the parameters from XLM-R as well as W jointly by maximising the log-probability of the correct label. The architecture diagram of the classification is shown in Figure \ref{fig:architecture}. We specially used the XLM-R large model.

\subsection{Transfer-learning strategies}
\label{subsec:transfer}
When we adopt XLM-R for multilingual offensive language identification, we perform transfer learning in two different ways. 

\paragraph{Inter-language transfer learning} We first trained the XLM-R classification model on first level of English offensive language identification dataset (OLID) \cite{OLID}. Then we save the weights of the XLM-R model as well as the softmax layer. We use this saved weights from English to initialise the weights for a new language. To explore this transfer learning aspect we experimented on Hindi language which was released for HASOC 2019 shared task \cite{hasoc2019}, on Spanish data released for Hateval 2019 \cite{hateval2019} and on Arabic, Danish, Greek and Turkish data releasred for OffensEval 2020 \cite{zampieri-etal-2020-semeval}.
  
\paragraph{Inter-task and inter-language transfer learning} Similar to inter-language transfer learning strategy, we first trained the XLM-R classification model on the first level of English offensive language identification dataset (OLID) \cite{OLID}. Then we only save the weights of the XLM-R model and use the saved weights to initialise the weights for a new language. We did not use the weights of the last softmax layer since we wanted to test this strategy on different data that has a different number of offensive classes to predict. Since the last softmax layer reflects the number of classes, it is not possible to transfer the weights of the softmax layer when the number of the classes are different in the classification task. We explored this transfer learning aspect with  Bengali dataset released with TRAC - 2 shared task \cite{trac2020}. As described in the Section \ref{sec:data} the classifier should make a 3-way classification in between ‘Overtly Aggressive’, ‘Covertly Aggressive’ and ‘Non Aggressive’ text data.

We used a Nvidia Tesla K80 GPU to train the models. We divided the data set into a training set and a validation set using 0.8:0.2 split on the dataset. We mainly fine tuned the learning rate and number of epochs of the classification model manually to obtain the best results for the validation set. We obtained $1e^{-5}$ as the best value for learning rate and 3 as the best value for number of epochs for all the languages. The other configurations of the transformer model were set to a constant value over all the languages in order to ensure consistency between the languages. This also provides a good starting configuration for researchers who intend to use this technique on a new language. We used a batch-size of eight, Adam optimiser \cite{kingma2014adam} and a linear learning rate warm-up over 10\% of the training data. The models were trained using only training data. We performed early stopping if the evaluation loss did not improve over ten evaluation rounds.

Training for English language took around 1 hour while training for other languages took around 30 minutes. 

\section{Results and Evaluation}
\label{sec:results}

We evaluate the results obtained by all models using the test set provided by the organizers of each competition. We compared our results to the best systems in TRAC-2 for Bengali, HASOC for Hindi, HatEval for Spanish in terms of weighted and macro F1 score according to the metrics reported by the task organizers - TRAC-2 reported only macro F1, HatEval reported only weighted F1 and HASOC reported both macro F1 and weighted F1. Finally, we quantify the improvement of the transfer learning strategy in the performance of both BERT and XLM-R. \textit{TL} indicates that the model used the transfer learning strategy described in Subsection \ref{subsec:transfer}. 

\begin{table*}[htb]
\centering
\scalebox{0.95}{
\begin{tabular}{l|ccc|ccc|ccc|c}

\hline
                                     & \multicolumn{3}{c|}{\textbf{Non Hate Offensive}} & \multicolumn{3}{c|}{\textbf{Hate Offensive}}             & \multicolumn{3}{c|}{\textbf{Weighted Average}}      & \textbf{}         \\ \hline
\multicolumn{1}{l|}{\textbf{Model}} & \textbf{P}   & \textbf{R}   & \textbf{F1}   & \textbf{P} & \textbf{R} & \textbf{F1}               & \textbf{P} & \textbf{R} & \textbf{F1}               & \textbf{F1 Macro} \\ \hline
\textit{XLM-R (TL)}                      & 0.84         & 0.87         & 0.88          & 0.84       & 0.86       & 0.85 & 0.86       & 0.85       & 0.86 & 0.86     \\
\textit{BERT-m (TL)}                     & 0.80         & 0.83         & 0.84          & 0.80       & 0.82       & 0.81 & 0.82       & 0.81       & 0.82 & 0.82              \\
\textit{\citet{bashar2019qutnocturnal}}  & NR           & NR           & NR            & NR         & NR         & NR   & NR         & NR         & 0.82 & 0.81              \\
\textit{XLM-R}                           & 0.84         & 0.96         & 0.90          & 0.86       & 0.61       & 0.71 & 0.85       & 0.85       & 0.81 & 0.80              \\
\textit{BERT-m}                          & 0.77         & 0.99         & 0.86          & 0.94       & 0.33       & 0.49 & 0.82       & 0.78       & 0.80 & 0.80              \\
\hline
\end{tabular}
}
\caption[Results for Hindi]{Results for offensive language detection in Hindi. For each model, Precision (P), Recall (R), and F1 are reported on all classes, and weighted averages. Macro-F1 is also listed. NR indicates that the research did not report that value. \citet{bashar2019qutnocturnal} is the best result submitted to the competition.}
\label{table:hindi}
\end{table*}

\begin{table*}[htb]
\centering
\scalebox{0.98}{
\begin{tabular}{l|ccc|ccc|ccc|c}

\hline
                                     & \multicolumn{3}{c|}{\textbf{Non Hateful}} & \multicolumn{3}{c|}{\textbf{Hateful}}             & \multicolumn{3}{c|}{\textbf{Weighted Average}}      & \textbf{}         \\ \hline
\multicolumn{1}{l|}{\textbf{Model}} & \textbf{P}   & \textbf{R}   & \textbf{F1}   & \textbf{P} & \textbf{R} & \textbf{F1}               & \textbf{P} & \textbf{R} & \textbf{F1}               & \textbf{F1 Macro} \\ \hline
\textit{XLM-R (TL)}                  & 0.78         & 0.76         & 0.77          & 0.76       & 0.78       & 0.73 & 0.76       & 0.76       & 0.76   & 0.75     \\
\textit{BERT-m (TL)}                 & 0.76         & 0.74         & 0.75          & 0.74       & 0.76       & 0.71 & 0.74       & 0.74       & 0.74   & 0.73              \\
\textit{\citet{vega2019mineriaunam}} & NR           & NR           & NR            & NR         & NR         & NR   & NR         & NR         & 0.73   & NR               \\
\textit{\citet{perez2019atalaya}}    & NR           & NR           & NR            & NR         & NR         & NR   & NR         & NR         & 0.73   & NR              \\
\textit{XLM-R}                       & 0.77         & 0.74         & 0.74          & 0.73       & 0.76       & 0.70 & 0.73       & 0.73       & 0.73   & 0.72     \\
\textit{BERT-m}                      & 0.75         & 0.72         & 0.73          & 0.71       & 0.74       & 0.68 & 0.72       & 0.72       & 0.72   & 0.71              \\
\hline
\end{tabular}
}
\caption[Results for Spanish]{Results for offensive language detection in Spanish. For each model, Precision (P), Recall (R), and F1 are reported on all classes, and weighted averages. Macro-F1 is also listed. NR indicates that the research did not report that value. \citet{vega2019mineriaunam} and \citet{perez2019atalaya} are the best results submitted to the competition.}
\label{table:spanish}
\end{table*}

\begin{table*}[htb]
\centering
\scalebox{0.92}{
\begin{tabular}{l|ccc|ccc|ccc|ccc|c}

\hline
                                     & \multicolumn{3}{c|}{\textbf{Non Aggressive}} & \multicolumn{3}{c|}{\textbf{Overtly Aggressive}}    & \multicolumn{3}{c|}{\textbf{Covertly Aggressive}}   & \multicolumn{3}{c|}{\textbf{Weighted Average}}      & \textbf{}         \\ \hline
\multicolumn{1}{l|}{\textbf{Model}} & \textbf{P}   & \textbf{R}   & \textbf{F1}   & \textbf{P} & \textbf{R} & \textbf{F1}   & \textbf{P} & \textbf{R} & \textbf{F1} & \textbf{P} & \textbf{R} & \textbf{F1}     & \textbf{F1 Macro} \\ \hline
\textit{XLM-R (TL)}                  & 0.96     & 0.94      & 0.95    & 0.73       & 0.81   & 0.77   & 0.72  & 0.80    & 0.76    & 0.86      & 0.85 & 0.84 & 0.84 \\
\textit{\citet{risch2020bagging}}    & NR       & NR        & NR      & NR         & NR     & NR     & NR    &  NR     & NR      & NR        & NR   & 0.82 & 0.82           \\
\textit{BERT-m (TL)}                 & 0.94     & 0.92      & 0.93    & 0.71       & 0.79   & 0.75   & 0.70  & 0.78    & 0.68    & 0.80      & 0.83 & 0.82 & 0.82           \\
\textit{XLM-R}                       & 0.91     & 0.90      & 0.91    & 0.70       & 0.78   & 0.74   & 0.69  & 0.77    & 0.67    & 0.79      & 0.83 & 0.82 & 0.81  \\
\textit{BERT-m}                      & 0.90     & 0.89      & 0.90    & 0.69       & 0.77   & 0.73   & 0.68  & 0.76    & 0.66    & 0.78      & 0.81 & 0.81 & 0.81             \\
\hline
\end{tabular}
}
\caption[Results for Bengali]{Results for offensive language detection in Bengali. For each model, Precision (P), Recall (R), and F1 are reported on all classes, and weighted averages. Macro-F1 is also listed. NR indicates that the research did not report that value. \citet{risch2020bagging} is the best result submitted to the competition.}
\label{table:bengali}
\end{table*}

For Hindi, as presentend in Table \ref{table:hindi}, transfer learning with XLM-R cross lingual embeddings provided the best results achieving 0.86 for both weighted and macro F1 score. In HASOC 2019 \cite{hasoc2019}, the best model by \citet{bashar2019qutnocturnal} scored 0.81 Macro F1 and 0.82 Weighted F1 using convolutional neural networks.  

For Spanish transfer learning with XLM-R cross lingual embeddings also provided the best results achieving 0.75 and 0.76 macro and weighted F1 score respectively. The best two models in HatEval \cite{hateval2019} for Spanish scored 0.73 macro F1 score.
Both models applied SVM classifiers trained on a variety of features like character and word n-grams, POS tags, offensive word lexica, and embeddings.

The results on Bengali shown in Table \ref{table:bengali} deserve special attention because the Bengali data is off-domain with respect to the English data (Facebook instead of Twitter) and it contains three labels (covertly aggressive, overtly aggressive, and not aggressive) instead of two in the English dataset (offensive and non-offensive). \textit{TL} indicates that the model used the inter-task, inter-domain, and inter-language transfer learning strategy described in Subsection \ref{subsec:transfer}. Similar to the Hindi and Spanish, for Bengali also transfer learning with XLM-R cross lingual embeddings provided the best results achieving 0.75 and 0.76 macro and weighted F1 respectively thus outperforming the other models by a significant margin. The best model in the TRAC-2 shared task \cite{trac2020} scored 0.82 weighted F1 score in Bengali using a BERT-based system.

\begin{figure}[ht]
\centering
\includegraphics[scale=0.60]{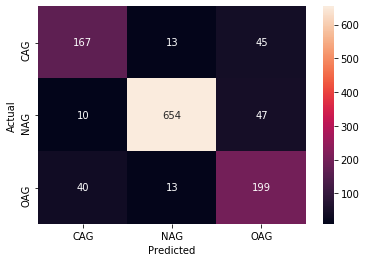}
\caption{Heat map of the Bengali test set predictions by XLM-R (TL).}
\label{fig:heatmap}
\end{figure}

We look closer to the test set predictions by XLM-R (TL) for Bengali in Figure \ref{fig:heatmap}. We observe that the performance for the non-aggressive class is substantially better than the performance for the overtly aggressive and covertly aggressive classes following a trend observed by the TRAC-2 participants including \citet{risch2020bagging}.

\noindent Finally, it is clear that in all the experimental settings, the cross-lingual embedding models fine-tuned with transfer learning, outperforms the best system available for the three languages. Furthermore, the results show that the cross-lingual nature of the XLM-R model provided a boost over the multilingual BERT model in all languages tested. 

\subsection{OffensEval 2020}
\label{offenseval}

We also experimented our methods with the recently released OffensEval 2020 \cite{zampieri-etal-2020-semeval} datasets in four languages: Arabic, Danish, Greek and Turkish. In Danish and Turkish we evaluate our results on the test sets comparing them to the best systems in the competition. As shown in Table \ref{table:danish} and Table \ref{table:greek} our methodology out performs the best systems submitted in each of these languages. As we mentioned earlier, remaining languages of OffensEval 2020 \cite{zampieri-etal-2020-semeval}: Arabic and Turkish are morphologically rich languages and they require language specific segmentation techniques which we hope to explore in future research. That said, in order to show our transfer learning strategy works for a wider range of languages, we presents the results we obtained for development set in the Tables \ref{table:arabic} and \ref{table:turkish}. Although not fully comparable, the high results obtained on the development set provides us an indication that the proposed approach is likely to obtain high results on the Arabic and Turkish test sets as well. We aim to perform this evaluation with the different morphological segmentation techniques in both Arabic and Turkish.

Finally, in all of the OffensEval 2020 languages, XLM-R with the transfer learning strategy performs best from the experiments. It is clear that transfer learning approach boosts the performance of both BERT-m and XLM-R models through out all the languages. Furthermore, it is clear that there is a significant advantage in using cross lingual transformers with transfer learning rather than training transformer models from scratch.

\begin{table*}[htb]
\centering
\scalebox{1}{
\begin{tabular}{l|ccc|ccc|ccc|c}

\hline
                                     & \multicolumn{3}{c|}{\textbf{Not Offensive}} & \multicolumn{3}{c|}{\textbf{Offensive}}             & \multicolumn{3}{c|}{\textbf{Weighted Average}}      & \textbf{}         \\ \hline
\multicolumn{1}{l|}{\textbf{Model}} & \textbf{P}   & \textbf{R}   & \textbf{F1}   & \textbf{P} & \textbf{R} & \textbf{F1}               & \textbf{P} & \textbf{R} & \textbf{F1}               & \textbf{F1 Macro} \\ \hline
\textit{XLM-R (TL)}                  & 0.85         & 0.88         & 0.89          & 0.69       & 0.75       & 0.72   & 0.87       & 0.87       & 0.87 & 0.83     \\
\textit{BERT-m (TL)}                 & 0.89         & 0.87         & 0.88          & 0.67       & 0.73       & 0.70   & 0.85       & 0.86       & 0.86 & 0.82              \\
\textit{\citet{offenseval2020ID14}}                 & NR        & NR         & NR          & NR       & NR       & NR   & NR       & NR       & NR & 0.81              \\
\textit{XLM-R}                       & 0.87         & 0.85         & 0.86          & 0.66       & 0.72       & 0.70   & 0.85       & 0.85       & 0.85 & 0.81       \\
\textit{BERT-m}                      & 0.86         & 0.84         & 0.85          & 0.65       & 0.71       & 0.69   & 0.84       & 0.84       & 0.84 & 0.80              \\
\hline
\end{tabular}
}
\caption[Results for Danish]{Results for offensive language detection in Danish. For each model, Precision (P), Recall (R), and F1 are reported on all classes, and weighted averages. Macro-F1 is also listed. NR indicates that the research did not report that value. \citet{offenseval2020ID14} is the best result submitted to the competition.}
\label{table:danish}
\end{table*}

\begin{table*}[htb]
\centering
\scalebox{1}{
\begin{tabular}{l|ccc|ccc|ccc|c}

\hline
                                     & \multicolumn{3}{c|}{\textbf{Not Offensive}} & \multicolumn{3}{c|}{\textbf{Offensive}}             & \multicolumn{3}{c|}{\textbf{Weighted Average}}      & \textbf{}         \\ \hline
\multicolumn{1}{l|}{\textbf{Model}} & \textbf{P}   & \textbf{R}   & \textbf{F1}   & \textbf{P} & \textbf{R} & \textbf{F1}               & \textbf{P} & \textbf{R} & \textbf{F1}               & \textbf{F1 Macro} \\ \hline
\textit{XLM-R (TL)}                  & 0.95         & 0.93         & 0.94          & 0.72       & 0.80       & 0.76   & 0.91       & 0.91       & 0.91 & 0.87     \\
\textit{BERT-m (TL)}                 & 0.94         & 0.92         & 0.93          & 0.71       & 0.79       & 0.75   & 0.90       & 0.90       & 0.90 & 0.86              \\
\textit{\citet{offenseval2020ID287}}                 & NR         & NR         & NR          & NR       & NR       & NR   & NR       & NR       & NR  & 0.85              \\
\textit{XLM-R}                       & 0.91         & 0.89         & 0.90          & 0.68       & 0.75       & 0.71   & 0.86       & 0.86       & 0.87 & 0.83     \\
\textit{BERT-m}                      & 0.90         & 0.88         & 0.89          & 0.67       & 0.74       & 0.70   & 0.85       & 0.85       & 0.85 & 0.81              \\
\hline
\end{tabular}
}
\caption[Results for Greek]{Results for offensive language detection in Greek. For each model, Precision (P), Recall (R), and F1 are reported on all classes, and weighted averages. Macro-F1 is also listed. NR indicates that the research did not report that value. \citet{offenseval2020ID287} is the best result submitted to the competition.}
\label{table:greek}
\end{table*}

\begin{table*}[htb]
\centering
\begin{tabular}{l|ccc|ccc|ccc|c}

\hline
                                     & \multicolumn{3}{c|}{\textbf{Not Offensive}} & \multicolumn{3}{c|}{\textbf{Offensive}}             & \multicolumn{3}{c|}{\textbf{Weighted Average}}      & \textbf{}         \\ \hline
\multicolumn{1}{l|}{\textbf{Model}} & \textbf{P}   & \textbf{R}   & \textbf{F1}   & \textbf{P} & \textbf{R} & \textbf{F1}               & \textbf{P} & \textbf{R} & \textbf{F1}               & \textbf{F1 Macro} \\ \hline

\textit{XLM-R (TL)}                  & 0.96         & 0.94         & 0.95          & 0.73       & 0.81       & 0.77 & 0.92       & 0.92       & 0.92 & 0.88     \\
\textit{BERT-m (TL)}                 & 0.95         & 0.92         & 0.93          & 0.72       & 0.78       & 0.76 & 0.90       & 0.90       & 0.90 & 0.86     \\
\textit{XLM-R}                       & 0.93         & 0.90         & 0.91          & 0.70       & 0.76       & 0.74 & 0.89       & 0.89       & 0.89 & 0.85     \\
\textit{BERT-m}                      & 0.91         & 0.89         & 0.90          & 0.70       & 0.75       & 0.49 & 0.87       & 0.87       & 0.88 & 0.84     \\
\hline
\end{tabular}
\caption[Results for Arabic]{Results for offensive language detection in Arabic. For each model, Precision (P), Recall (R), and F1 are reported on all classes, and weighted averages. Macro-F1 is also listed. NR indicates that the research did not report that value. \citet{offenseval2020ID145} is the best result submitted to the competition. }
\label{table:arabic}
\end{table*}

\begin{table*}[htb]
\centering
\begin{tabular}{l|ccc|ccc|ccc|c}

\hline
                                     & \multicolumn{3}{c|}{\textbf{Not Offensive}} & \multicolumn{3}{c|}{\textbf{Offensive}}             & \multicolumn{3}{c|}{\textbf{Weighted Average}}      & \textbf{}         \\ \hline
\multicolumn{1}{l|}{\textbf{Model}} & \textbf{P}   & \textbf{R}   & \textbf{F1}   & \textbf{P} & \textbf{R} & \textbf{F1}               & \textbf{P} & \textbf{R} & \textbf{F1}               & \textbf{F1 Macro} \\ \hline
\textit{XLM-R (TL)}                  & 0.83         & 0.81         & 0.82          & 0.70       & 0.78       & 0.74 & 0.89       & 0.89       & 0.89 & 0.85     \\
\textit{BERT-m (TL)}                 & 0.81         & 0.79         & 0.80          & 0.69       & 0.77       & 0.72 & 0.87       & 0.88       & 0.88 & 0.84              \\
\textit{XLM-R}                       & 0.79         & 0.77         & 0.78          & 0.67       & 0.75       & 0.70 & 0.85       & 0.85       & 0.86 & 0.80     \\
\textit{BERT-m}                      & 0.77         & 0.75         & 0.77          & 0.66       & 0.74       & 0.69 & 0.84       & 0.84       & 0.85 & 0.79              \\
\hline
\end{tabular}
\caption[Results for Turkish]{Results for offensive language detection in Turkish. For each model, Precision (P), Recall (R), and F1 are reported on all classes, and weighted averages. Macro-F1 is also listed. NR indicates that the research did not report that value. \citet{offenseval2020ID228} is the best result submitted to the competition.}
\label{table:turkish}
\end{table*}

\subsection{Progress Tests}\begin{figure}
\centering
  \begin{subfigure}[b]{7cm}
    \centering\includegraphics[width=7cm]{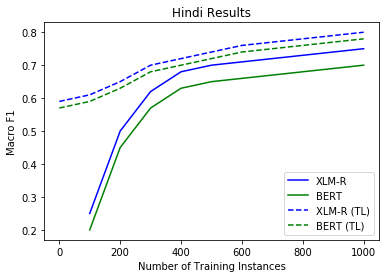}
    \caption{Hindi Results}
    \label{fig:hindi_results}
  \end{subfigure}
  \begin{subfigure}[b]{7cm}
    \centering\includegraphics[width=7cm]{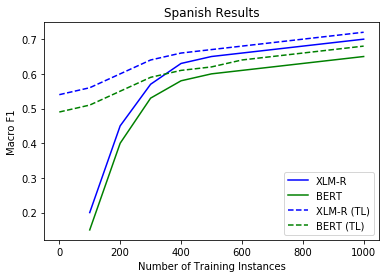}
    \caption{Spanish Results}
    \label{fig:spnaish_Results}
  \end{subfigure}
  \begin{subfigure}[b]{7cm}
    \centering\includegraphics[width=7cm]{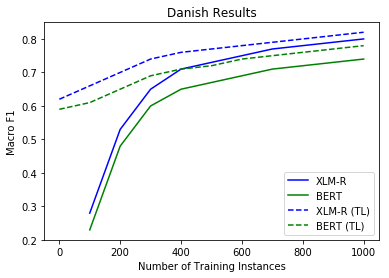}
    \caption{Danish Results}
    \label{fig:danish_Results}
  \end{subfigure}
  \begin{subfigure}[b]{7cm}
    \centering\includegraphics[width=7cm]{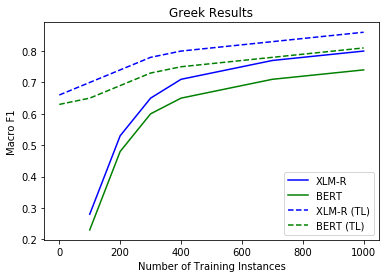}
    \caption{Greek Results}
    \label{fig:greek_Results}
  \end{subfigure}
  \begin{subfigure}[b]{7cm}
    \centering\includegraphics[width=7cm]{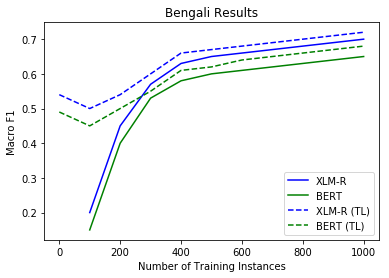}
    \caption{Bengali Results}
    \label{fig:bengali_Results}
  \end{subfigure}
 
\caption{Transfer learning impact in Offensive Language Identification. \textbf{XLM-R} and \textbf{BERT-m} indicates that the model was trained from scratch while \textbf{XLM-R (TL)} and \textbf{BERT (TL)} indicates that models followed the transfer learning strategy.}
\label{fig:progress_results}
\end{figure}

The biggest challenge in developing deep learning models for offensive language identification in low resources languages is finding suitable annotated data. Therefore, we investigate whether our transfer learning approach can thrive in even less resource environments. In order to analyze this, we conduct the experiments for 0 (unsupervised), 100, 200, 300 and up to 1,000 training instances. We performed this in five languages: Hindi, Spanish, Danish, Greek and Bengali. We compare the results with and without transfer learning for both BERT-m and XLM-R. We report the Macro-F1 score of the predictions and gold labels in the test set against the number of training instances in Figure \ref{fig:progress_results}.

The transfer learning strategy significantly impacts the results. For Hindi, with only 100 training instances, training XLM-R from scratch achieves only 0.25 Macro F1 between the predictions and gold labels of the test set. However, with only 100 training instances, training XLM-R using the transfer learning strategy achieves 0.61 Macro F1 in Hindi. When the number of training instances grows, the results from the XLM-R models trained with the transfer learning strategy and the results from the XLM-R models trained from scratch converge. Similar pattern can be observed with BERT-m model too. This can be noticed in all the languages we experimented. However, in Bengali increasing number of training instances from 0 to 100 dropped the transfer learning results a bit, most probably because it takes certain number of examples to train the softmax layer we added.

It is interesting to see that, in all the languages, transfer learning strategy with zero number of training instances provided a good Macro F1 score. Therefore, it is clear that the cross lingual nature of these transformer models can identify offensive content in an unsupervised environment too, once we do transfer learning from English.

Our results confirm that XLM-R with the transfer learning strategy can be hugely beneficial to low-resource languages in offensive language identification where annotated training instances are scarce.  

\section{Conclusion and Future Work}
\label{sec:future}

This paper is the first study to apply cross-lingual contextual word embeddings in offensive language identification projecting predictions from English to other languages. We developed systems based on different transformer architectures and evaluated their performance on benchmark datasets on Bengali \cite{trac2020}, Hindi \cite{hasoc2019}, and Spanish \cite{hateval2019}. Finally, we compared them with the performance obtained by the systems that participated in shared tasks and we have showed that our best models outperformed the best systems in these competitions. Furthermore, our methods obtained promising results on the Arabic, Danish, Greek, and Turkish training and development sets released as part of the recent OffensEval 2020 \cite{zampieri-etal-2020-semeval}.

We have showed that XLM-R with transfer learning outperfoms all other methods we tested as well as the best results obtained by participants of the three competitions. Furthermore, the results from Bengali show that it is possible to achieve high performance using transfer learning on off-domain and off-task data when the labels do not have a direct correspondence in the projected dataset (two in English and three in Bengali). This opens exciting new avenues for future research considering the multitude of phenomena (e.g. hate speech, abuse, cyberbulling), annotation schemes, and guidelines used in offensive language datasets.

We are keen to cover more languages for offensive language identification using this strategy. Furthermore, when we are experimenting with morphologically rich languages like Arabic and Turkish we are interested to see whether language specific preprocessing like segmentation would improve the results. Finally, we would also like to apply our models to more low-resource languages creating important resources to cope with the problem of offensive language in social media. 

We are currently experimenting with the recently finished HASOC 2020 shared task on identifying offensive YouTube comments in Code-mixed Malayalam. We are keen to experiment our transfer learning strategy in this task since the dataset is off domain and it contains code-mixed data. To the best of our knowledge, this is the first time that such methods are being tested on code-mixed data. Initial experiments \cite{ranasinghe2020} show that XLM-R with transfer learning strategy provides 0.85 and 0.73 weighted F1 and macro F1 scores respectively while training XLM-R from scratch gets only 0.83 and 0.71 weighted F1 and macro F1 scores respectively. These results suggest that our transfer learning approach is very likely to perform well with code-mixed data compared to other state-of-the-art methods. 


\section*{Acknowledgments}

We would like to thank the shared task organizers for making the datasets used in this paper available. We further thank the anonymous reviewers who provided us with constructive and insightful feedback to improve the quality of this paper.


\bibliographystyle{ACM-Reference-Format}
\bibliography{EMNLP2020}

\end{document}